\documentclass{article}

     \usepackage[final]{neurips_2022}

\usepackage[utf8]{inputenc} 
\usepackage[T1]{fontenc}    
\usepackage{hyperref}       
\usepackage{url}            
\usepackage{booktabs}       
\usepackage{amsfonts}       
\usepackage{nicefrac}       
\usepackage{microtype}      
\usepackage{xcolor}         

\usepackage{amsmath,amsfonts,bm}

\def\eqref#1{equation~\ref{#1}}

\def\1{\bm{1}}

\DeclareMathAlphabet{\mathsfit}{\encodingdefault}{\sfdefault}{m}{sl}
\SetMathAlphabet{\mathsfit}{bold}{\encodingdefault}{\sfdefault}{bx}{n}

\newcommand{\Set}[1]{\mathcal{#1}}

\usepackage{graphicx}
\usepackage{caption}
\usepackage{subcaption}
\usepackage{comment}

\usepackage{makecell}
\usepackage{wrapfig}

\def\abovestrut#1{\rule[0in]{0in}{#1}\ignorespaces}

\newcommand{\smallgap}{\abovestrut{0.15in}}

\title{Foundation Models for Semantic Novelty in Reinforcement Learning}

\newcommand{\specificthanks}[1]{\@fnsymbol{#1}}

\author{%
Tarun Gupta{\normalfont \textsuperscript{1,2}}\thanks{Correspondence to \href{mailto:tarun.gupta@cs.ox.ac.uk}{tarun.gupta@cs.ox.ac.uk}. Tarun has done this work as an intern at NVIDIA Research. }$\;\;$
Peter Karkus{\normalfont \textsuperscript{1}}\thanks{Equal contribution.}$\;\;$
Tong Che{\normalfont \textsuperscript{1}}\footnotemark[2]$\;\;$
Danfei Xu{\normalfont \textsuperscript{1,3}}$\;\;$
Marco Pavone{\normalfont \textsuperscript{1,4}}$\;\;$
\\
  \textsuperscript{1}NVIDIA Research $\;\;$
  \textsuperscript{2}University of Oxford$\;\;$ 
  \textsuperscript{3}Georgia Institute of Technology$\;\;$ 
  \textsuperscript{4}Stanford University $\;\;$
}

\begin{document}

\maketitle

\begin{abstract}
    Effectively exploring the environment is a key challenge in reinforcement learning (RL). 
    We address this challenge by defining a novel intrinsic reward based on a foundation model, such as contrastive language image pretraining (CLIP),  which can encode a wealth of domain-independent semantic visual-language knowledge about the world. 
    Specifically, our intrinsic reward is defined based on pre-trained CLIP embeddings without any fine-tuning or learning on the target RL task. We demonstrate that CLIP-based intrinsic rewards can drive exploration towards semantically meaningful states and outperform state-of-the-art methods in challenging sparse-reward procedurally-generated environments.
\end{abstract}

\section{Introduction}

Exploration is a key challenge in reinforcement learning (RL), especially when extrinsic feedback is sparsely provided by the environment. Intrinsic motivation  provides an effective way for agents to  explore the environment by rewarding the visitation of novel states \citep{oudeyer2009intrinsic, aubret2019survey}. 
An important question is then the definition of state novelty.
Most existing methods compute novelty scores based on some form of distance function in a learned state representation, which is typically learned simultaneously with the policy \citep{bellemare2016unifying, pathak2017curiosity}.
For example, a popular recent approach RIDE \citep{raileanu2020ride} defines novelty as the distance of consecutive states in the embedding space of a dynamics model learned alongside the policy. 


In this paper we propose a \emph{foundation model} based intrinsic motivation scheme for RL, FoMoRL. 
Instead of computing novelty scores based on a domain-specific learned state representations, FoMoRL uses the pre-trained state representation of a foundation model, i.e., a generic model trained on large domain-independent visual-language data, such as contrastive language image pretraining (CLIP)  \citep{radford2021learning}.
In this manner FoMoRL may incorporate semantic knowledge about the state space without requiring any domain-specific pre-training, and thus help guide exploration towards semantically meaningful states such as picking up keys, opening the doors, and interacting with other useful objects relevant to the task.

The idea of FoMoRL is general to the choice of RL algorithm,  motivation scheme, and the choice of foundation model. We apply FoMoRL on top of  RIDE \citep{raileanu2020ride} and the IMPALA distributed off-policy RL algorithm \citep{espeholt2018impala}; and use CLIP as our foundation model \citep{radford2021learning}. 
Formally, FoMoRL can be applied to any visual RL tasks, with or without language conditioning; however, FoMoRL is expected to be effective in domains where human-interpretable semantic knowledge is useful to succeed. 


We evaluate FoMoRL on the challenging sparse-reward procedurally-generated MiniGrid domain \citep{minigrid}. 
Procedurally generated environments present a significant learning challenge as an agent is unlikely to visit a state more than once, and therefore the learning method needs to generalize to unseen scenarios. Somewhat surprisingly, our results suggest that CLIP representations can be very effective even in grid-world environments, despite that  CLIP was trained on general human language-vision data that contains little (if any) grid-world like data points. Our analysis shows that CLIP representations encode fine-grained discrimination between different objects, colors, counts, and other visual attributes, which makes them effective for novelty based exploration even in grid-worlds. FoMoRL significantly outperforms state-of-the-art RIDE \citep{raileanu2020ride} method in 6 out of 9 challenging MiniGrid exploration tasks, and performs similarly in the rest of the tasks.  FoMoRL learns to solve the tasks 20\% faster, and achieve over 86\% higher average return per episode across all tasks.




\section{Related Work}
\textbf{Exploration in RL}: There are many techniques for efficient exploration in model-free single-agent RL, including but not limited to intrinsic novelty reward \citep{bellemare2016count, tang2017hash}, predictability \citep{pathak2017curiosity, oudeyer2007intrinsic, stadie2015incentivizing}, pure curiosity \citep{burda2019curiosity}, Bayesian posteriors \citep{osband2016rvf, gal2017dropout, fortunato18noisynet, odonoghue2018ube}, information gain \citep{houthooft2016vime} or empowerment \citep{klyubin2005empowerment, choshen2018dora}. 

For intrinsic motivation-based exploration, a key problem is defining the novelty score. \citet{raileanu2020ride} proposed RIDE which computes novelty scores based on differences between state representations of consecutive states. The state representations are learned from learning a transition function via forward and inverse dynamics model. Another work on random network distillation (RND) \citep{burda2018exploration} uses a fixed randomly initialized neural network to represent the states and reward the exploration based on the error of a neural network predicting the fixed features. Never Give Up (NGU; \citep{badia2020never}) combines an episodic novelty module based on inverse dynamics features with RND as a lifelong novelty module. DIAYN \citep{eysenbach2018diversity} performs unsupervised exploration independent of the task and models the policies conditioned on a latent skill variable. In contrast to these works we propose to use pre-trained foundation models to define intrinsic rewards.

Concurrent to our efforts, \citet{tam2022semantic} proposed the idea of using pretrained representations from foundation models for exploration in RL. The high-level idea in the concurrent work is the same as ours; however, there are important differences in its application. The concurrent work builds on top of RND \citep{burda2018exploration} and NGU \citep{badia2020never} whereas we use RIDE \citep{raileanu2020ride}; and the concurrent work mostly focuses on Unity based 3D environments with language oracles, whereas we experiment with procedurally-generated long-horizon grid-world tasks that have no language oracle, and whose visual appearance differs significantly from the visual data used to pretrain the CLIP model.

\textbf{Foundation Models in RL}: 
\citet{mu2022improving} uses language to improve exploration via intrinsic rewards instead of using raw states, however their method requires a oracle language annotator which is not easily available for many RL environments. \citet{khandelwal2022simple} investigate the effectiveness of CLIP visual representations directly for control on Embodied AI tasks \citep{batra2020objectnav} by bypassing the learning of policy representations with CLIP embeddings. Their results demonstrated the effectiveness of CLIP representations for control on navigation-heavy Embodied AI tasks. 
\citet{shridhar2022cliport} proposed CLIPort which uses CLIP representations with imitation learning for robotic manipulation tasks, adding spatial understanding of the scenes as well. SayCan \citep{ahn2022can} uses a Large Language Model (LLM) to supply high-level instructions to solve the tasks and pair it with an affordance based value function to see if the robot has required skills to perform those instructions. Similarly another paper \citep{patel2021mapping} shows that LLMs such as GPT-2 \citep{radford2019language}  or GPT-3 \citep{brown2020language} can learn to ground the concepts such as direction or colour that it is explicitly taught and also generalise to several instances of unseen concepts as well. 
In contrast to the above works we explore using intrinsic rewards based on foundational models for efficient semantic exploration in visual RL tasks.

\section{Problem Formulation}
A partially-observable Markov decision process (POMDP) is defined as a tuple $G=\left\langle \Set S,\Set A, P, R, \Omega, O, \gamma \right\rangle$. The true state of the environment is denoted by $s \in \Set S$. At each time step, an agent chooses an action $a \in \Set A$, which causes a transition in the environment according to the state transition kernel $P(s'|s,a):\Set S\times\Set A\times \Set S \rightarrow [0,1]$, and the agent receives an extrinsic reward  $R_{\text{ext}}: \Set S\times\Set A\rightarrow\mathbb{R}$. $\gamma\in[0,1)$ is a discount factor.

Due to \textit{partial observability}, the agent cannot observe the true state $s$, but receives a partially observable visual image $o \in \Omega$ drawn from observation kernel $o \sim O(s,a)$. At time $t$, the agent has access to its action-observation history $\tau_t \in \Set T_t \equiv (\Omega \times \Set U)^t \times \Omega$, on which it conditions a stochastic policy $\pi(a_t|\tau_t)$. The stochastic policy induces an action-value function : $Q^\pi(s_t, \tau_t, a_t)=\mathbb{E} \left[G_t|s_t,\tau_t,a_t\right]$, where $G_t=\sum^{\infty}_{i=0}\gamma^ir_{t+i}$ is the \textit{discounted return}.

\section{FoMoRL}
FoMoRL assigns an intrinsic reward based on CLIP's \citep{radford2021learning} visual encoder embeddings. CLIP embeddings enable semantic understanding of the images as it has been contrastively trained with image captioned large scale human data, thereby incorporating language abstractions. These language abstractions can summarise the images with important fine grained details and yet being very concise, much like our natural languages \citep{borghi2014words}.

\subsection{Semantic Intrinsic Motivation}
We define the representation of the current observation $\phi(o_t)$ as the pretrained representation from the CLIP encoder i.e. $\phi(o_t) = \text{clip}(o_t)$. The intrinsic reward $R_{\text{int}}$ at time step $t$ is computed as the L2-norm $||\phi(o_{t+1}) - \phi(o_{t}) ||_2$ of the difference in the clip representation between consecutive states discounted by episodic state visitation counts similar to \citet{raileanu2020ride}:
\begin{equation*}
    r_{\text{int}}(s_t,a_t,o_t) = \frac{||\text{clip}(o_{t+1}) - \text{clip}(o_{t}) ||_2}{\sqrt{N_{ep}({s_{t+1})}}},
\end{equation*}
where $N_{ep}(s_{t+1})$ is the visitation count of state $s_{t+1}$ during the current episode, which is initialized to 1 in the beginning of the episode. The state visitation count can be directly computed for MDPs with small number of discrete states, but can also be approximated for MDPs with large discrete or continuous state spaces \citep{martin2017count, machado2020count}. The above intrinsic reward encourages the visitation of states which have significantly different CLIP representations from the current state. CLIP representations can perform fine-grained classification of abstract concepts in the visual observations and can therefore guide exploration towards semantically meaningful states which are useful for the downstream task. The division by state counts ensures that the agent does not exploit the intrinsic reward function by swinging between two states with large difference in CLIP embeddings. For this paper, we use the officially released ResNet$50\times16$ model for the CLIP encoder which outputs state representations of dimension $1024$. 

\subsection{Training Setup}
We train our policy using IMPALA \citep{espeholt2018impala} in a distributed off-policy setup with multiple actors. The policy is trained to maximize the combined reward i.e. $r(s_t, a_t) = \alpha r_i(s_t, a_t) + r_e(s_t, a_t)$, where $\alpha$ is a hyperparameter. The policy learns its own state representations from scratch, while the pretrained representations are just used for guiding exploration through intrinsic rewards. The total loss $L_{\text{tot}}$ function is composed of four components: $L_{\text{pg}}$: policy gradient loss, $L_{\text{base}}$: value function loss, and $L_{\text{ent}}$: entropy loss, with respective hyper parameter coefficients i.e.
\begin{equation*}
    L_{\text{tot}} = L_{\text{pg}} + c_{\text{base}} L_{\text{base}} + c_{\text{ent}} L_{\text{ent}}
\end{equation*}

Since our policy is conditioned on partial observations, we use an LSTM \citep{hochreiter1997long} to summarize the action-observation history $\tau_t$. 

\section{Experiments}
\subsection{MiniGrid Domain}
\label{domainexp}
We evaluate our proposed method on three types of hard exploration tasks from the MiniGrid domain \citep{minigrid}: MultiRoomN$X$S$Y$, KeyCorridorS3R3, and ObstructedMaze2Dlh. The environment consists of an $N \times M$ grid world with each cell is either empty or consist one of the objects such as wall, floor, lava, door, key, ball, box and goal. The agent can pick up and carry exactly one object, and each object has a discrete color. In order to open up a locked door, the agent has to pick up and carry the key matching the color of the door. The action space is discrete and allows agent to move forward, turn left/right, pick up an object, drop the object being carried, toggle (open doors or interact with objects), and a done action to mark the task complete. At each time step, the agent receives a partial and egocentric observation and the agent cannot see through walls or closed doors.

\begin{figure}
     \centering
     \begin{subfigure}[b]{0.3\textwidth}
         \centering
         \includegraphics[width=\textwidth]{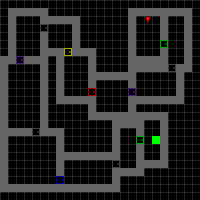}
         \caption{MultiRoomN12S10}
         \label{fig:MGA}
     \end{subfigure}
     \hfill
     \begin{subfigure}[b]{0.3\textwidth}
         \centering
         \includegraphics[width=\textwidth]{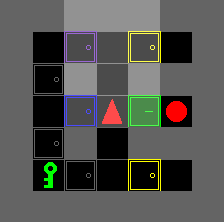}
         \caption{KeyCorridorS3R3}
         \label{fig:MGB}
     \end{subfigure}
     \hfill
     \begin{subfigure}[b]{0.3\textwidth}
         \centering
         \includegraphics[width=\textwidth]{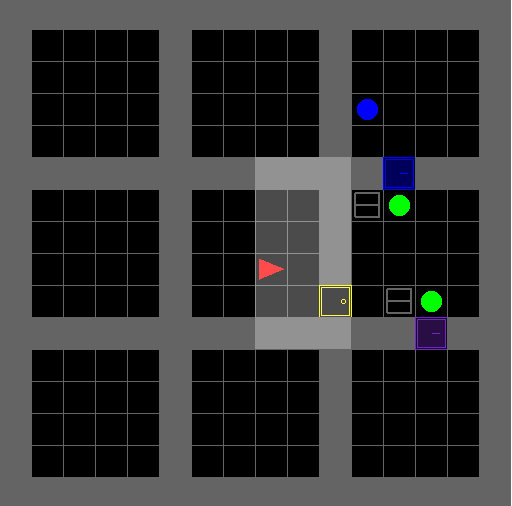}
         \caption{ObstructedMaze2Dlh}
         \label{fig:MGC}
     \end{subfigure}
        \caption{Representative examples of different MiniGrid environments. The highlighted area showcases the partial observations the agent receives.}
        \label{fig:three graphs}
\end{figure}

The MultiRoomN$X$S$Y$ task (Figure \ref{fig:MGA}) consists of $X$ rooms, with size at most $Y$, connected in random orientations in every episode. This task has a series of connected rooms with doors that must be opened in order to get to the next room with the green goal square as the final destination the agent must get to. In the KeyCorridorS3R3 task (Figure \ref{fig:MGB}), the agent has to explore the environment to find a key hidden in another room, and then use the key to open the corresponding door and pickup the ball. In the ObstructedMaze2Dlh task (Figure \ref{fig:MGC}), the agent has to pick up a box which is placed in a corner of a maze. The doors are locked, the keys are hidden in boxes and doors are obstructed by balls. The NoisyTV \citep{burda2018exploration} version of MultiRoomN$X$S$Y$ adds stochasticity to the environment by adding a random ball which changes color everytime the agent takes a particular action. The agents need to learn to avoid this distraction as it is irrelevant to the task.

The environment provides different input representations of the state: (1) compact grid encoding using 3 integer values: describing the type and color of the object in the cell, and a flag indicating whether doors are open or closed; (2) raw RGB visual observations from the procedurally generated environment.

\subsection{Baseline}
For all our experiments, we use IMPALA \citep{espeholt2018impala} as the base RL algorithm, and compare against the popular RIDE method by \citet{raileanu2020ride}. RIDE also uses IMPALA to train the policy with the intrinsic reward based on learning state representations which minimizes the forward and inverse dynamics loss. In the original paper, the authors use the compact grid encoded representation (see Section \ref{domainexp} for definition), which already contains all key required information about the visual observation, thereby making it simpler to learn a transition function and the state representations. In this paper, we instead use the raw visual RGB observations, which makes the task significantly harder as agents need to extract the features from raw images. Both the policy and intrinsic reward modules for RIDE and FoMoRL method receives a partial egocentric observation as input.

\subsection{Results}

\begin{table*}
\centering
\caption{Performance of RIDE and FoMoRL on a variety of hard exploration problems in MiniGrid. Both methods receive partially observable RGB images as input to the policy and intrinsic reward module. The values indicate the final mean episodic return averaged over multiple seeds after training on [$x$M] frames. The bracketed values ($x$M) indicate the average number of training frames (in millions) required to achieve the convergence value. The (-) indicates no convergence in [$x$M] frames. }
\vspace{1ex}
\begin{tabular}{lcc}
\toprule
& \makecell{RIDE} & \makecell{FoMoRL} \\
\midrule
DoorKey-5x5 [15M] & \textbf{0.96} (5M) & \textbf{0.96 (2.5M)}\\
\smallgap
MultiRoom-N7-S4 [30M]  & 0.71 (15M) & \textbf{0.76 (7M)}\\
MultiRoomNoisyTV-N7-S4 [30M]  & 0.7 (20M) & \textbf{0.74 (19M)}\\
MultiRoom-N7-S8 [30M] & 0.0 (-) & \textbf{0.59 (29M)}\\
MultiRoom-N10-S4 [30M] & 0.66 (10M) & \textbf{0.75 (6M)} \\
MultiRoom-N10-S10 [80M] & 0.0 (-) & 0.0 (-)\\
MultiRoom-N12-S10 [80M] & 0.0 (-) & 0.0 (-)\\
\smallgap
ObstructedMaze-2Dlh [80M] & 0.0 (-) & \textbf{0.96 (45M)}\\
KeyCorridorS3R3 [30M] & 0.0 (-) & \textbf{0.9 (18M)}\\
\bottomrule
\end{tabular}
\label{tab:scores}
\end{table*}

\begin{table*}
\centering
\caption{Performance of FoMoRL on very-hard exploration tasks N10-S10 and N12-S10 in MiniGrid with partially observable RGB images as input to the policy and fully observable RGB images as input to the intrinsic reward module. The values indicate the final mean episodic return averaged over multiple seeds after training on [$x$M] frames. The bracketed values ($x$M) indicate the average number of training frames (in millions) required to achieve the convergence value. }
\vspace{1ex}
\begin{tabular}{lc}
\toprule
& \makecell{FoMoRL} \\
\midrule
MultiRoom-N10-S10 [80M] & \textbf{0.63 (22M)} \\
MultiRoom-N12-S10 [80M] & \textbf{0.64 (24M)}\\
\bottomrule
\end{tabular}
\label{tab:scoresfull}
\end{table*}
Table \ref{tab:scores} summarizes the results for various MiniGrid tasks. As shown in the table, FoMoRL consistently outperforms the original baseline method RIDE both with respect to better final average return and time to reach convergence. RIDE fails to learn efficient state representations in the partially observable RGB space and therefore completely fails to learn in a number of scenarios. On the other hand, FoMoRL directly uses CLIP representations as state embeddings for intrinsic motivation and thereby completely bypass the representation learning altogether. 

As shown in the table, both RIDE and FoMoRL failed to learn anything on larger versions of MultiRoom environment i.e. MultiRoom-N10-S10 and MultiRoom-N12-S10 with 10 and 12 rooms respectively having a maximum room size of 10. We analyzed the training logs of these failed tasks and found that the failure is attributed to the partial observability of the RGB input image. Since intrinsic reward module only sees a partial observable egocentric observation, it tries to generate diversity within that restricted view and gather intrinsic reward. Therefore, the agent learns to move around from one cell to another and open doors only when it is within the observable area. The moving around from cell to cell ends up taking most of the horizon time before the agent can reach the final destination. We therefore tried to resolve this by  allowing only the intrinsic module to access the full RGB image of the environment, and keeping the policy input partially observable. Table \ref{tab:scoresfull} clearly shows that providing the intrinsic reward module access to fully observable RGB images resolves the issue for FoMoRL.

For some tasks, we noticed that higher value of intrinsic reward coefficient led to the agent getting distracted from solving the original task and rather focus on generating more and more diversity, thereby making both RIDE and FoMoRL fail to learn. However, most environments were solvable by $\alpha$=\{0.1, 0.5\} for RIDE and $\alpha$=\{0.02, 0.005\} for FoMoRL.

\subsection{CLIP Analysis on MiniGrid}

\begin{wrapfigure}{r}{0.35\textwidth}
  \begin{center}
    \includegraphics[width=0.3\textwidth]{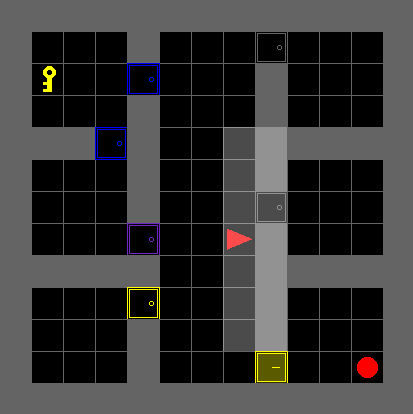}
  \end{center}
  \caption{RGB Input Image for CLIP Embedding Analysis}
  \label{clip-test}
\end{wrapfigure}

\begin{figure}
     \centering
     \begin{subfigure}[b]{0.8\textwidth}
         \centering
         \includegraphics[width=\textwidth]{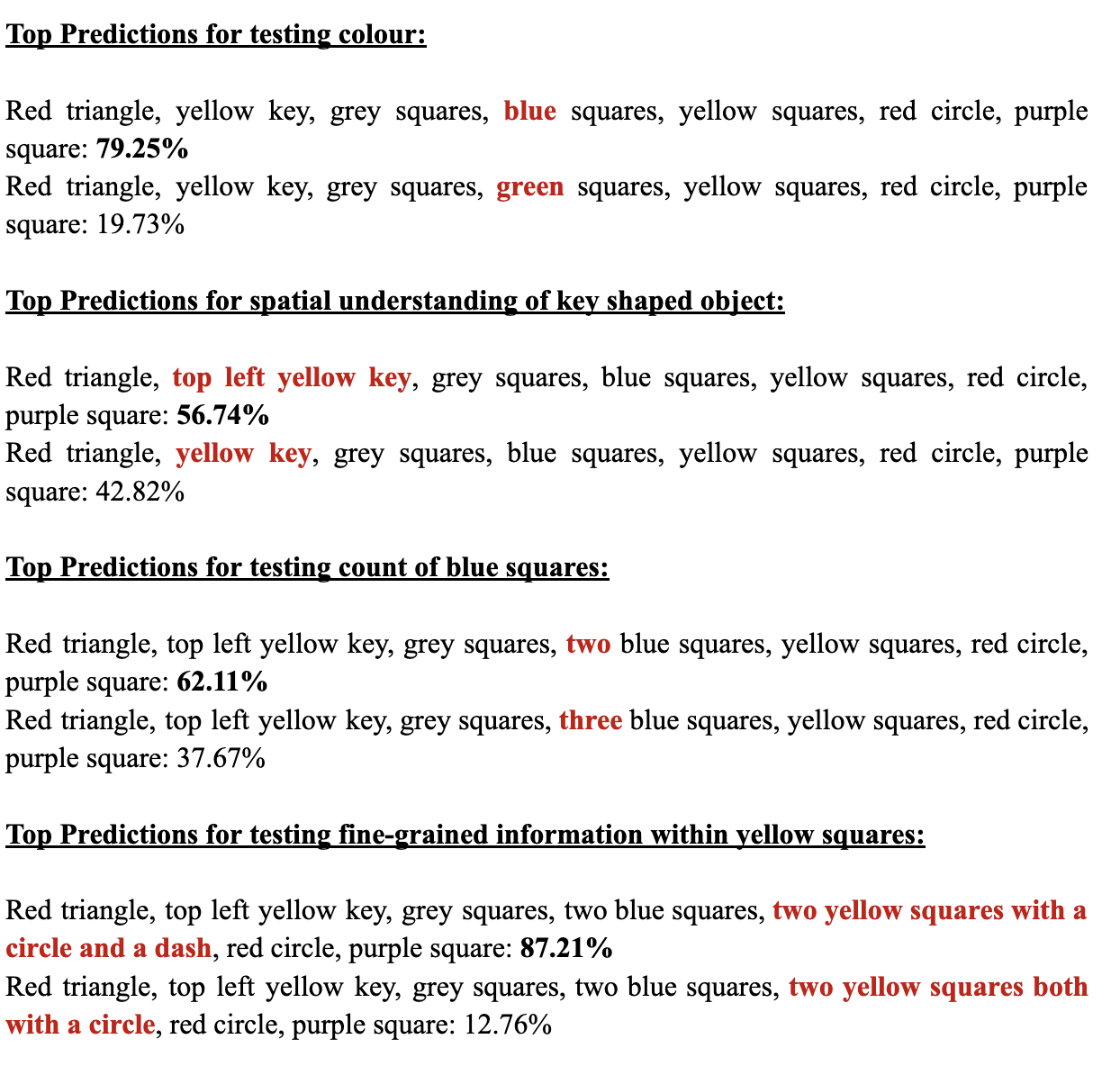}
     \end{subfigure}
    \caption{Similarity scores for proposed language descriptions of the input test image. (a) Test to check whether CLIP representations prefer a description with the presence of a green square (incorrect) or a grey square (correct) in the test image. (b) Test to check the count of blue squares in the test image. (c) Test to check if CLIP representations prefer the description with spatial information of the key in the the test image or not. (d) Test to check more fine-grained understanding of the test image by comparing against the shapes present within the yellow square in the test image.}
        \label{fig:analysis}
\end{figure}

In the previous section, we showed that CLIP representations are remarkably effective across multiple MiniGrid exploration tasks. We perform a small experiment to check what exactly CLIP understands from these RGB grid images. We input the RGB images to the CLIP's visual encoder and then input proposed language descriptions through the CLIP's language encoder and compute a similarity score. The higher score of similarity implies that CLIP representations prefer that description over others. Figure \ref{clip-test} shows the test image which is inputted to the CLIP's visual encoder. The proposed language descriptions and the similarity scores are shown in \ref{fig:analysis}.

Figure \ref{fig:analysis} shows the results for testing the CLIP representations to encode the color and count of objects, their spatial information, and more fine-grained information. We test the color understanding of the CLIP representations by comparing a correct description of the color of squares with an incorrect one, as there are no green squares present in the test image (Figure \ref{clip-test}). We then inspect if CLIP representations prefers a description with spatial information of the key shaped object or not. To check the understanding of CLIP representations to represent the count of objects, we compare the number of blue squares with an incorrect description of three blue squares.  Lastly, we test for more fine grained information within the yellow squares. One of the yellow squares has a circle within it and other one has a dash within (semantically meaning that the door is unlocked/locked). 

Figure \ref{fig:analysis} indicates that CLIP representations can clearly discriminate between different colors and counts of objects, their spatial understanding, and even more fine grained information like whether a square has a mini circle or minus within. CLIP representations does not directly understand the significance of the objects within the RGB image, for example, it does not directly understand that a square indicates a door, or a square with a mini circle indicates an unlocked door. However, the representations can understand and distinguish between the geometrical shapes very well which is enough for diversity based semantic exploration. On the contrary, CLIP representations have already shown to be very effective for large range of tasks from text based video retrieval \citep{fang2021clip2video, luo2021clip4clip}, text driven image manipulation \citep{patashnik2021styleclip} to embodied AI tasks \citep{khandelwal2022simple} based on realistic visual observations, for example, a kitchen scene containing a microwave. 

\section{Conclusion}
We showcase that language abstractions incorporated within foundation models such as CLIP can be very useful for semantic exploration in RL, even on grid based environments. Our proposed method FoMoRL can be applied independent of the choice of RL algorithm, intrinsic motivation scheme, or the type of foundation model. FoMoRL achieves state-of-the-art performance with significantly higher average return and solving the task much faster than the baseline RIDE method. This work shows a proof of concept for leveraging foundation models for semantic novelty in RL. Future work may investigate realistic 3D environments like Habitat ObjectNav Challenge \citep{batra2020objectnav} and language conditioned tasks such as CALVIN \citep{mees2022calvin}. It would be also interesting to leverage the language encoder from CLIP alongside the visual encoder to enable semantic exploration in language conditioned tasks.

\bibliographystyle{plainnat}
\bibliography{neurips_2022}

\end{document}